\pdfoutput=1
\documentclass[11pt]{article}

\usepackage{acl}

\usepackage{times}
\usepackage{latexsym}
\usepackage{array}  % For table
\usepackage{multirow}  % For appendix table
\usepackage{multicol}
\usepackage{booktabs}
\usepackage{listings}

\usepackage{pgfplots}
\pgfplotsset{compat=1.18}
\usepgfplotslibrary{polar}
\usepgfplotslibrary{fillbetween} 
\usepackage{pgfplotstable}

\usepackage{algorithm}
\usepackage{algorithmic}

\usepackage{booktabs}
\usepackage{colortbl}
\usepackage{adjustbox}
\usepackage{graphicx}

\usepackage[T1]{fontenc}
\usepackage[utf8]{inputenc}

\usepackage{microtype}

\usepackage{inconsolata}

\usepackage{graphicx}
\usepackage{colortbl}
\usepackage{booktabs}
\usepackage{adjustbox}
\usepackage{graphicx}
\definecolor{gold}{HTML}{ffeb80}
\definecolor{silver}{HTML}{ece2e2}
\definecolor{bronze}{HTML}{f9c593}

\usepackage[inline]{enumitem} % For inline listing
\usepackage{amsmath}
\usepackage{tcolorbox}
\usepackage{tikz}
\usetikzlibrary{positioning}
\usepackage{float}
\title{Are You Sure You’re Positive? Consolidating Chain-of-Thought Agents with Uncertainty Quantification for Aspect-Category Sentiment Analysis}

\author{
  \textbf{Filippos Ventirozos}\textsuperscript{1,2},
  Peter Appleby\textsuperscript{2},
  Matthew Shardlow\textsuperscript{1}
  \\
  \\
  \textsuperscript{1}Manchester Metropolitan University, \\
  \textsuperscript{2}Autotrader Research Group, Autotrader UK \\
  \\
  \small{
    \textbf{Correspondence:} \href{mailto:f.ventirozos@mmu.ac.uk}{f.ventirozos@mmu.ac.uk}
  }
}

\begin{document}
\maketitle
\begin{abstract}

% version 5
Aspect-category sentiment analysis provides granular insights by identifying specific themes within product reviews that are associated with particular opinions. Supervised learning approaches dominate the field. However, data is scarce and expensive to annotate for new domains. We argue that leveraging large language models in a zero-shot setting is beneficial where the time and resources required for dataset annotation are limited. Furthermore, annotation bias may lead to strong results using supervised methods but transfer poorly to new domains in contexts that lack annotations and demand reproducibility. In our work, we propose novel techniques that combine multiple chain-of-thought agents by leveraging large language models’ token-level uncertainty scores. We experiment with the 3B and 70B+ parameter size variants of Llama and Qwen models, demonstrating how these approaches can fulfil practical needs and opening a discussion on how to gauge accuracy in label-scarce conditions.

\end{abstract}

% TL;DR:

\section{Introduction}
Tracking customer satisfaction is critical for organisations aiming to improve their products and services. However, traditional supervised approaches require bespoke datasets, necessitating time and human labour \citep{Pustejovsky_Stubbs_2013}. The costs of annotation can be a significant challenge for many organisations, prohibiting access to adapted state-of-the-art solutions, and even to evaluation. Hence, in this paper, we explore and compare two popular large language models (LLMs) across various low-training scenarios, specifically zero-shot, for analysing customer satisfaction. 

Sentiment analysis approaches are commonly used to classify feedback as positive, neutral, or negative for customer satisfaction, but this method often lacks granularity. A single piece of feedback can express multiple sentiments targeting different product or service aspects, leading to ambiguous classifications and limiting actionable insights. To address this, aspect-based sentiment analysis (ABSA) methodologies \citep{9996141} have been developed to more accurately capture the nuanced sentiments present in customer feedback.

In our study, we investigated the ABSA task of aspect category sentiment analysis (ACSA), a type of ABSA. For a given piece of text, ACSA considers the aspect categories and sentiment polarities, as shown in Figure~\ref{fig:quad_example}. The categories are a set of classes which are predefined by domain experts or stakeholders. Characteristically, in the figure, the example demonstrates two tuples from the same text. Firstly, the review found the pepperoni pizza, a type of \#Food (Category), to be delicious, representing a positive sentiment (Polarity). Secondly, the review found the service to be terrible, representing negative polarity.

\begin{figure}[h]
    \centering

\begin{tcolorbox}[colframe=black, colback=gray!10, width=2.6in, title=ACSA Tuple Example]
\textbf{Input:} \textit{The pepperoni pizza was delicious but the service was terrible though.}
\\~\\
\textbf{Output:}
\begin{minipage}[t]{0.55\linewidth}
  \centering
  \vspace{-\baselineskip} % Adjust vertical alignment
  \begin{tabular}{|c|c|}
    \hline
    \textbf{Category} & \textbf{Polarity} \\
    \hline
    \#Food & positive \\
    \hline
    \#Service & negative \\
    \hline
  \end{tabular}
\end{minipage}
\end{tcolorbox}

    \caption{An ACSA example where a review results in two category-sentiment polarity pairs.}
    \label{fig:quad_example}
\end{figure}

Supervised learning approaches hold the state of the art for ACSA tasks \citep{cai-etal-2020-aspect,PING2024126994,Xu2025}. In contrast, we explore the zero-shot setting, which eliminates the need for labelled instances and allows us to experiment with LLM agents on novel datasets. This approach is particularly compelling in real-world applications, as organisations and firms often lack the resources or time required to annotate large datasets.

In this paper, we propose a novel approach that integrates multiple LLM agents, each utilising a distinct chain-of-thought (CoT) reasoning process. Prior research \citep{fei-etal-2023-reasoning,10499502} has introduced a sequential CoT approach, where aspects are first identified, followed by opinions and then polarities. In our experiments, we adapt this framework to the ACSA task and critically evaluate the assumptions underlying CoT by exploring alternative reasoning sequences. Additionally, we introduce mechanisms to combine the outputs of each CoT agent through aggregation techniques that make use of the LLMs' token confidence scores, also referred to as token-level uncertainty.

% For our experiments we utilised Llama and Qwen LLM and we compared them with providing in-context samples as noted from the literature.

The primary contributions of this study are enumerated below:
\begin{enumerate}[itemsep=0.2pt]
  \item We experimented with how the ordering of CoT prompt elements affects LLMs' performance across datasets.

\item We compared multiple aggregation methods for effectively combining outputs from multiple CoT agents.

\item We benchmark our approach against a previously published zero-shot method, reproduced and evaluated on new datasets.

%\item We present novel findings and an in-depth discussion regarding using token log probability scores as confidence measures for each LLM output.

% \item We analysed and compared performance and confidence scores across two popular open-source LLMs.
\end{enumerate}

\section{Related Work}
\label{sec:RW}

ABSA has emerged as a particularly interesting research topic due to its increasing popularity and widespread applicability across various domains \citealp{rink-etal-2024-aspect,Namee2022,10099815,arianto-budi-2020-aspect,Chu2022,10151883} inter alia).

% In this paper, we tackled the ACSA task a sub-task of ABSA, which is particularly relevant for industries since the ACSA task involves pairing category terms with sentiment polarities, which can be readily quantified and analysed (see Section~\ref{sec:cot_agents}).

In this paper, we address the ACSA task, a type of ABSA. The ACSA task is especially relevant for industry applications, as it involves associating predefined category terms with corresponding sentiment polarities, which can be readily quantified and analysed (more on Section~\ref{sec:cot_agents}).

Although a few studies have considered the ACSA task from a supervised learning perspective \citep{cai-etal-2020-aspect,PING2024126994,Xu2025}, it would be beneficial to address it in an unsupervised manner because annotation requires considerable time and human labour, and may also obfuscate reproducibility, as discussed later in this paper.

Zero-shot and few-shot methodologies have been employed for the broader ABSA task. For instance, \citet{Hellwig2025DoWS} use LLMs with a few-shot in-context learning (ICL) approach, while \citet{10499502} evaluate different prompting techniques for sentiment analysis using LLMs. \citet{fei-etal-2023-reasoning} investigated a multi-hop CoT approach, and \citet{bai-etal-2024-compound} introduced ChatABSA, an ICL few-shot prompt framework for ABSA that also extends to the ACSA task we address.

In our paper, we aim to push the boundaries of zero-shot learning by examining the use of multiple LLM agents provided with no prior knowledge. More agents can yield better results; indeed, \citet{li2024agentsneed} discuss how employing additional agents and subsequently performing majority voting can improve outcomes.

To the best of our knowledge, there is no prior work on multi-agent collaboration for ACSA. We therefore experiment with this approach, inspired by previous ABSA research on CoT \citep{fei-etal-2023-reasoning,10499502}, and we investigate different aggregation techniques that leverage token-level uncertainty in LLMs \citep{shorinwa2024surveyuncertaintyquantificationlarge}. 

\section{Methodology}

\subsection{Problem Statement}
Our problem statement aligns with prior work on ACSA. Given a text (whether a single sentence from a review or an entire review) our goal is to extract one or more pairs, each consisting of a category and its corresponding polarity, as depicted in Figure \ref{fig:quad_example}. While the order of the pairs may vary due to the generative nature of the models used, the internal structure of each pair must remain consistent: the category always precedes the sentiment polarity. This ensures clarity and uniformity in the extracted results. Such as:

\begin{equation}
    Q = \{(c_i, p_i)\}_{i=1}^n
\end{equation}

where \(Q\) represents the set of pairs, \(c_i\) is the category, and \(p_i\) is the polarity for the \(i\)-th pair, and \(n\) is the number of pairs extracted from the text. In the following subsections, we describe the different methods utilised in our experiments.

% \subsection{In-Context Learning}
% We utilised ICL techniques similarly to \citet{Xu2023TheLO} and \citet{Zhang2023SentimentAI}. Specifically, in a multi-turn setting, we provided demonstrations on how the LLM should parse a sentence into a quadruple.

% We ensured that there were as many samples as possible for each combination of category and polarity throughout our ICL examples. If the number of examples was less than the number of categories, we opted to include each sample based on a different category.

% \subsubsection{Prompt Crafting}
% Firstly, following the recommendations of \citet{10499502}, we employed a role-playing instruction for the LLM, designating it as an NLP assistant expert in ABSA. This approach required the LLM to provide precise answers and strictly adhere to the given instructions. 

% Subsequently, we adopted the prompt style of \citet{Xu2023TheLO} and \citet{Zhang2023SentimentAI}. We indicated whether an element must be extracted from the text or could be left \textquotesingle NULL\textquotesingle. We then listed the possible categories of the domain, specified the polarity range, and showcased the format of a quadruple. Based on the number of demonstrations, we supplied a history of conversations to be used as context for the LLMs. Figure \ref{fig:icl_example} shows an example with one demonstration.

\subsection{CoT Agents}
\label{sec:cot_agents}
Previous research on related ABSA tasks \citep{fei-etal-2023-reasoning,10499502} have typically followed an intuitive sequence: first extracting aspects, and then identifying opinions and their corresponding polarities. However, in our experiments, we challenge this conventional approach by exploring alternative reasoning pathways. Specifically, we design and evaluate multiple LLM agents, each guided by a distinct prompt that dictates a unique CoT process for extracting category-polarity pairs.

To better understand the ACSA task, let us break it down step by step. In a review, aspect terms are the words or phrases that refer to specific features of a product or service, such as ``pepperoni pizza'' in our example in Figure~\ref{fig:quad_example}. While these aspect terms are helpful, they are not always ideal for quantitative analysis. For instance, rather than analysing sentiment for individual dishes (e.g., pepperoni pizza, pasta) that may vary across restaurants, it is more meaningful to map these terms to broader categories, such as ``Food'', to measure overall sentiment on food for each restaurant. Similarly, opinions like ``delicious'' provide qualitative insights but are not easily quantifiable. Mapping these opinions to sentiment polarities—positive, neutral, or negative—enables more actionable and comparative analytics.

In our CoT experiments, we prompted the LLMs to detect one element at a time, ultimately generating category-polarity pairs. We systematically permutated the order of the three elements: aspect terms, aspect categories, and opinions. Sentiment polarity was always derived from the opinion generated at the end of the prompt. For example, one agent might follow the sequence $aspects \rightarrow categories \rightarrow opinions$, while another might use $opinions \rightarrow aspects \rightarrow categories$. For simplicity, we refer to the entities in the CoT sequences (i.e., aspects, categories, and opinions) as ``elements''. This systematic variation of element ordering enables us to investigate how reasoning sequences affect the accuracy of sentiment analysis prediction.

 % Table \ref{} outlines the different CoT sequences used in our prompts for each agent.
% In our preliminary experiments, we observed no significant performance gains when explicitly including opinions as an intermediate step (see Appendix~\ref{}). 

\subsubsection{Prompt Crafting}
Our next step involved designing the necessary prompts to chain the reasoning elements effectively. Previous research has explored multi-hop CoT approaches \citet{fei-etal-2023-reasoning,10499502}, where the process involves using a prompt to request each element from the LLM sequentially with a new call. However, in our preliminary experiments (see Appendix~\ref{sec:appendix:cot_vs_multi_hop}), we found that this multi-hop approach did not yield superior results compared to consolidating all instructions into a single prompt providing enumerated instructions. Consequently, we adopted an enumerated CoT reasoning process within a single prompt. Examples of our prompts are illustrated in Appendix~\ref{sec:appendix:cot_prompts}, and our multi-hop CoT prompts adjusted for our experiments are in Appendix~\ref{sec:appendix:multi_hop}.

The enumerated instructions inside the prompt were manually crafted in an imperative language, inspired by previous research in the space \citep{fei-etal-2023-reasoning,10499502,bai-etal-2024-compound}. To ensure that our final prompts are reasonably robust, we evaluated them on the inter-prompt stability score \citep{promptstability}, using the 3B Llama model as a reference.  We noticed our prompt maintained a constant inter-prompt stability score across different ranges of temperatures (0.1, 0.5, 1.0, 2.0, 5.0) for paraphrasing the prompt with the encoder-decoder PEGASUS model \citep{pegasus}, where stability across temperatures for paraphrasing is optimal, whereas if it deviates, is not \citep{promptstability}.

% mentioned it was designed for binary outputs mostly so their rough estimate 0.8 may not entice us completly

% Additionally, as guidance, we used the inter-prompt stability score \citep{promptstability} to evaluate our prompts for being robust for a range of the default temperatures and across our datasets, where our final prompt reached an inter-pss score of 0.74.

% . We then tested their inter-prompt stability score \citep{promptstability} to make sure our prompts were robust, we acquired an average score of 0.79 across the smaller LLM models, for all the datasets and for temperature sizes of 0.1 to 5.0 with .1 increments, meaning that they were fairly robust.

In the last part of the prompt, we instructed the LLM to format the output as Python code, specifically as a list of pairs. Each pair consists of two elements: the category and the sentiment polarity. Additionally, we incorporated a system instruction, following the approach of prior research \citep{10499502}, which served as a role-playing mechanism. This instruction constrained the LLM to generate the most probable answers only, minimising verbosity. 

% Appendix~\ref{sec:appendix:cot_prompts} provides in more detail the prompts and shows an example of a CoT prompt along with the system instruction. 

\subsubsection{Post-Processing}
From the LLM's text-generated response, we would parse the output string into Python code, neglecting any text generated outside the list. Then, we would use the native's Pyhon diff lib library\footnote{The \href{https://github.com/python/cpython/blob/main/Lib/difflib.py}{difflib} library works by finding the longest continuous characters between strings.} to map the category from the generated text to the list of potential categories. This was done to avoid discrepancies and spelling mistakes in generation when classifying the category-polarity pairs.

\subsection{CoT Aggregation Techniques}
\label{sec:aggregation_techniques}
After we obtained each agent's results, we explored various techniques and algorithms for aggregating the resulting pairs of each agent. For most of our algorithms, we would rely on the token-level uncertainty produced by the LLMs. 

\begin{figure*}[h]
\centering
    \includegraphics[width=0.6\textwidth]{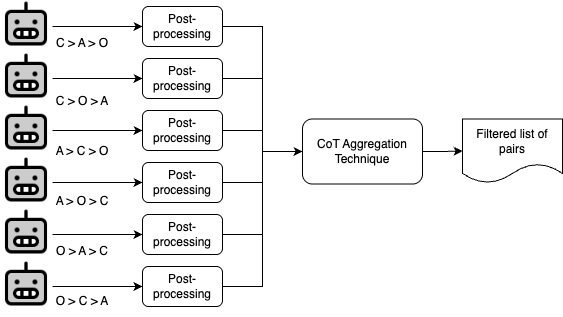}
    \caption{On the left side, we present the various CoT agents derived from the same LLM, each employing a different CoT extraction order. The letters C, A, and O denote Category, Aspect, and Opinion respectively, while the chevrons between them indicate their extraction sequence. The outputs from these agents are subsequently post-processed and passed through an aggregation technique to obtain the final set of ACSA pairs.}
    \label{fig:workflow}
\end{figure*}

Specifically, during generation, the LLMs assign each token a logarithmic probability based on the conditional distribution given all preceding tokens. For convenience, we convert these logarithmic values into probabilities. Subsequently, we extract the category and sentiment polarity words for each pair, omitting any special characters (i.e. the Python brackets), and compute the average of the probabilities of these words to obtain the final probability score for each pair, also referred to as the pair's confidence score. Below we list the algorithms, where a list refers to a list of pairs generated by the language models:
\begin{enumerate}
    \item \textbf{Highest probability list} Given that each agent has generated a list of pairs, we would acquire the top list in confidence score from the pool of agents, based on the averaged pair probabilities for a given list.

    % where confidence is defined as the average log prob between all the generated category and sentiment-polarity tokens generated.
    \item \textbf{Most common list} We acquire the most common list of pairs between agents. If there is no majority, then pick the agent(s) pairs with the highest confidence score.
    
    \item \textbf{Highest probability pairs} We get the top $n$ most probable pairs across all lists (i.e. across all agents). We also evaluated different techniques for setting the $n$ for each prediction. More on that can be found in the Appendix \ref{sec:appendix:alpha}. 
    % For brevity on our results we will demonstrate the highest recorded score from the different $n$ settings.
    \item \textbf{Clustered pairs} After obtaining the number of pairs $n$, identically to the above algorithm, we cluster all the unique pairs from all the agents based on their category word embeddings into $n$ clusters using the k-means algorithm from the Scikit-learn library \citep{scikit-learn}. These embeddings are generated by the RoBERTa encoder model \citep{roberta}, fine-tuned in a semantic textual similarity sentence setting \citep{reimers-gurevych-2019-sentence}. Then, we select the pair with the highest probability score from each cluster. We chose a sentence embedding model as opposed to a word embedding one since one category can be multiple tokens, for instance, the category ``performance\#sizing/fit'' from the Shoes dataset.
    
    \item \textbf{Most confident agent} We select the agent whose predictions yield the highest sum of list confidence across the dataset, where the list confidence is again the average pair probabilities in that list.
    % \item \textbf{LLM as an Aggregator} We would feed in all the lists from all the agents in a same type LLM model, and then the LLM would choose the best tuples. More on the Appendix~ref{}.
\end{enumerate}
In Figure~\ref{fig:workflow} we show the workflow from the various CoT agents to funnelling through an aggregation technique to get the end list of pairs.

\section{Experiments}
\subsection{Datasets}

For our experiments, we included four datasets. We opted for the Laptop16 \citep{pontiki-etal-2016-semeval}, Restaurant16 \citep{pontiki-etal-2016-semeval}, MAMS \citep{jiang-etal-2019-challenge} and Shoes \citep{peper-etal-2024-shoes}. All of these datasets were developed for the ACSA task, apart from the Shoes one, which was developed with an aspect-sentiment quad prediction task (Aspect, Category, Opinion, Sentiment) in mind, which we repurposed for ACSA (using only the Category and Sentiment labels). 

Large-scale LLMs are known to be trained on existing code repositories and the open internet, which may include pre-existing datasets \citep{samuel-etal-2025-towards}. We purposely included the recent Shoes dataset since the LLM models do not exhibit any knowledge of this dataset, making it an ideal case to analyse performance on the totally ``unseen'' dataset. Moreover, the Shoes dataset is the only one which considers whole reviews as input, whereas the other datasets have segmented the reviews into sentences. In Appendix~\ref{sec:appendix:dataset_statistics}, Table \ref{tab:datasets}, we provide the number of instances for each data fold and the number of categories.

% \begin{table*}[h]
% \centering
% \begin{tabular}{lllll}
% \hline
% Datasets        & Laptop 16 & Restaurant 16 & MAMS & Shoes \\ \hline
% N. train samples     &  2468  & 1954 & 3149 & 906   \\
% N. val samples       &  n/a   & n/a & 400 & 116    \\
% N. test samples      &  579   & 571 & 400 & 125   \\
% N. categories        &  67    & 12  & 8   & 21  \\ \hline
% \end{tabular}
% \caption{Dataset statistics for the four datasets employed in our study. The number of instances comes after pre-processing, omitting any examples with conflicting labels. Laptop 16 and Restaurant 16 do not have a validation dataset.}
% \label{tab:datasets}
% \end{table*}

% In Appendix~\ref{sec:appendix:dataset_knowledge}, we showcase how the employed LLMs respond to knowledge awareness queries regarding the datasets we have used in our work. 

\subsection{LLM Models}
% For our experiments, we used the 72B and 3B Qwen model \citep{qwen2025qwen25technicalreport} and Llama 3B and 72B model \citep{grattafiori2024llama3herdmodels}. Since we want to compare the probabilities between tokens we opted for the greedy search decoding strategy. The detailed versions of our models are listed in Listing 1.

% {\small
% \begin{lstlisting}[language=, caption={LM versions used in our experiments:}, label=lst:llm_versions]
%    - Qwen/Qwen2.5-72B-Instruct
%    - Qwen/Qwen2.5-3B-Instruct
%    - meta-llama/Llama-3.3-70B-Instruct
%    - meta-llama/Llama-3.2-3B-Instruct
% \end{lstlisting}
% }

For our experiments, we used the 72B and 3B Qwen model \citep{qwen2025qwen25technicalreport} and Llama 3B and 72B model \citep{grattafiori2024llama3herdmodels}. Since we want to compare the probabilities between tokens, we opted for the greedy search decoding strategy. The detailed versions of our models are listed in \autoref{lst:llm_versions}.

{\small
\begin{lstlisting}[language=, caption={LM versions used in our experiments.}, label=lst:llm_versions, captionpos=b]
   - Qwen/Qwen2.5-72B-Instruct
   - Qwen/Qwen2.5-3B-Instruct
   - meta-llama/Llama-3.3-70B-Instruct
   - meta-llama/Llama-3.2-3B-Instruct
\end{lstlisting}
}

\subsection{Evaluation}

The ACSA task focuses solely on extracting the category and polarity tuples. This task is best characterised as a multi-label classification problem, as the categories are predefined per domain and the polarity values range across positive, neutral, and negative. We adhere to the same micro-F1 metric as those used in the previously mentioned evaluations, as disclosed in \citet{cai-etal-2020-aspect}.

% \begin{table}[H]
% \centering
% \begin{tabular}{|l|l|}
% \hline
% Category & Polarity \\ \hline
% \end{tabular}
% \end{table}

\subsection{Baseline}
% For the best CoT agents derived from the aforementioned feature elimination we tested how would they perform if we added on the prompt the same ICL samples with the ChatABSA prompt. We found from our results is that the CoT chains with ten examples performed better on average than ICL with ten examples. One hypothesis is that the ICL technique captures the annotator's bias towards labelling word spans, while for more objective labelling tasks, the CoT performs better. 
For benchmarking, we utilise the ChatABSA method \citep{bai-etal-2024-compound}. The authors of ChatABSA have designed a prompt specifically for the ACSA task, allowing the inclusion of a certain number of ICL samples as few-shot demonstrations. In our experiments, we adopt the zero-sample setting, as our goal is to perform comparisons in a zero-shot scenario.

% \paragraph{Recommended approach}
% The best approach would involve having the highest F1 score with the least amount of LLM queries. Hence, we believe that the best approach would be to use the 3 CoT+ICL highest prob aggregate (not multi-hop).

% \subsection{Benchmarks}
% According to the literature most emphasis is encoder-decoder ...

% We also adapted the work of ... THOR for comparison.

\section{Results \& Discussion}

\paragraph{Element Order Debunking}
We tested three element permutations for each LLM model (i.e., aspect, category and opinion). Our experiments on the four datasets demonstrated that the order of the three elements between the two larger LLM models was fairly consistent, but it differed for each dataset. For instance, order 1) Opinion 2) Category 3) Aspect produced higher F1 scores for the Laptop16 dataset, whereas order 1) Category 2) Aspect 3) Opinion was best for the MAMS dataset. For the other two datasets, although they differed, the position of the Opinion and the Categories always followed the same pattern, indicating a CoT trend per dataset.

When comparing the 3B LLMs, we notice the same pattern in that the CoT element order stays fairly consistent across each dataset. Moreover, for all the LLMs, we did not notice any CoT agent that consistently underperformed; similarly, it varied across datasets but not across the models. Thus, we draw two conclusions from our experiments. Firstly, as opposed to previous research \citep{fei-etal-2023-reasoning,10499502}, starting from the aspects in the CoT prompts does not always provide optimal results. Secondly, the order is rather data-dependent and stays consistent across our four models. In the case of our multi-hop CoT preliminary experiments, the best-performing agents always had the Opinion preceding the Category and Aspect, see Table~\ref{tab:cot_vs_multi_hop}. For more detailed results across the agents per model and dataset, one can look in the Appendix~\ref{sec:appendix:cot_agent_performance}.

% A peculiar find was that although intuitively retrieving the aspects first and then the categories in a CoT scenario would perform better in all cases apart, for the implementation on Llama for the shoe dataset, first extracting the Categories and then the aspect terms scored higher.

% \begin{figure}[h]
%     \centering
%     \includegraphics[width=1\columnwidth]{figures/393C048F-BF1F-4242-B0F8-8E4442D5021D_1_201_a.jpeg}
%     \caption{The performance of ASQP by different CoT agents across datasets and models. The x-axis labels show the sequence of CoT. For instance, extracting the sentiments first, then the categories, and finally the aspects would be represented as 1. Sent. 2. Cat. 3. Asp.}
%     \label{fig:cot_seq}
% \end{figure}

% \paragraph{Is multi-hop worth?}
% Our experiments demonstrated that using the multi-hop technique did not yield significantly better results; in some cases, the results were even worse.

\begin{figure*}[h]
    \centering
    \includegraphics[width=0.8\linewidth]{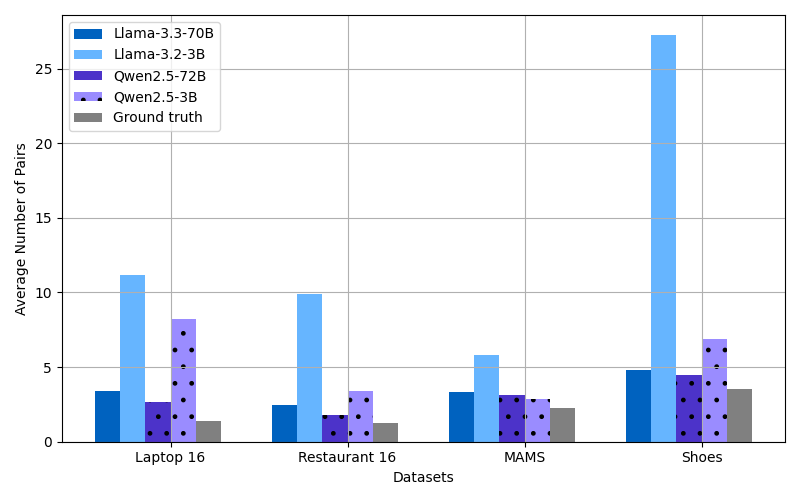}
    \caption{The average number of pairs generated for each model for each dataset. These refer to the \textit{Joined CoT Agent}, the pairs are summed. In dark grey is the average number of the ground truth for each dataset.}
    \label{fig:n_labels}
\end{figure*}

\paragraph{Joined CoT Agent}
We combined all pairs generated by the six CoT agents and refer to this combination as the \textit{joined CoT agent}. The results indicate that, as expected, precision decreased, but recall was higher than that of any individual CoT agent. Notably, for our models with 70B+ parameters, the average recall improvement was approximately 20\% across all datasets, except for the Restaurant16 dataset, since in that particular dataset, prediction scores were already relatively high. The 3B models did not substantially lag behind in recall scores with their larger counterparts. These findings illustrate the potential benefit of leveraging multiple agents from different CoT chains to enhance performance in ABSA tasks.

To better understand the performance of each LLM, we looked at all the aggregated pairs. It was evident that the smaller language models produced many more pairs, many of which were irrelevant. Especially the 3B Llama produced in one dataset 6x the number of pairs than its bigger 70B sibling, as seen in Figure~\ref{fig:n_labels}. Moreover, we counted the number of pairs where the category was the same, but there was more than one sentiment polarity with the same category. The 3B models, with Llama leading, demonstrated many more conflicting categories, whereas the larger ones were at a minimum, suggesting that the smaller LLMs struggled to delineate the polarities. One can view the number with conflicting pairs in Appendix~\ref{sec:appendix:joined}.

% In Appendix~\ref{sec:appendix:joined}, one can view the graph with the average number of pairs of the joined CoT agent for each dataset and the graph with the number of conflicting sentiment polarities. 

\paragraph{CoT Aggregation Techniques}

Subsequently, we compared the various aggregation techniques, as enumerated in Section~\ref{sec:aggregation_techniques}. The average results are presented in Table~\ref{tab:top7performers_percent}. It can be observed from the table that the \textit{highest probability list} provided the best overall results amongst all the aggregation techniques. These results suggest that the confidence provided by the LLMs can be useful. 

Following this, the \textit{most common tuples} among agents scored second, suggesting that relying on confidence scores is better than majority voting. Unfortunately, selecting pairs apart from the lists was not the most effective strategy: any strategy in this category—\textit{highest probability pairs} and \textit{clustered pairs} with any $n$ technique—underperformed compared to the two strategies that left the lists intact, although they appeared to achieve higher scores in the 3B model space. We speculate that this is because LLMs inherently can approximate the ground truth list of pairs, but they may deviate due to different interpretations that stray from the ground truth.

\begin{table*}[htbp]
\centering
\resizebox{\textwidth}{!}{
\begin{tabular}{lcccc}
\toprule
\textbf{LLM Configuration} & \textbf{Qwen2.5-72B-Instruct} & \textbf{Qwen2.5-3B-Instruct} & \textbf{Llama-3.3-70B-Instruct} & \textbf{Llama-3.2-3B-Instruct} \\
\midrule
ChatABSA fs:0 & 48.5\%  & 27.3\%  & 50.4\%  & 10.5\%  \\
\midrule
Highest prob. list & \cellcolor{silver}\textsuperscript{\tiny 2}55.5\%  & \cellcolor{gold}\textsuperscript{\tiny 1}36.7\%  & \cellcolor{silver}\textsuperscript{\tiny 2}54.3\%  & 23.4\% \\
Most conf. agent & 54.2\%  & \cellcolor{bronze}\textsuperscript{\tiny 3}35.0\% & 53.2\%  & 28.5\%  \\
Most common. list & \cellcolor{bronze}\textsuperscript{\tiny 3}54.4\%  & 34.0\%  & \cellcolor{bronze}\textsuperscript{\tiny 3}53.9\%  & \cellcolor{gold}\textsuperscript{\tiny 1}29.7\%  \\
Highest prob. pairs alpha:1 & 52.9\%  & 33.0\%  & 50.5\%  & 26.3\%  \\
Highest prob. pairs alpha:0.9 & 52.9\% & 32.8\%  & 50.5\%  & \cellcolor{bronze}\textsuperscript{\tiny 3}26.8\% \\
Highest prob. pairs alpha:mean & 51.9\%  & 31.0\%  & 49.7\%  & 25.6\%  \\
Highest prob. pairs alpha:max & 51.4\%  & \cellcolor{silver}\textsuperscript{\tiny 2}34.4\%  & 50.0\%  & \cellcolor{silver}\textsuperscript{\tiny 2}27.2\%  \\
\midrule
Top performant agent & \cellcolor{gold}\textsuperscript{\tiny 1}56.6\%  & 36.0\%  & \cellcolor{gold}\textsuperscript{\tiny 1}54.6\%  & 26.8\%  \\
\bottomrule
\end{tabular}}
\caption{The table demonstrates the F1 scores averages in percentages for each model averaged on our four datasets. The first one, Chat-ABSA, in the zero-shot setting, is our baseline. Following are our top 7 aggregation techniques, and lastly is the CoT agent, which scored the highest on each dataset, and then averaged. The top three F1 scores for each model are coloured in podium finish and annotated with superscripts (\textsuperscript{\tiny 1}gold, \textsuperscript{\tiny 2}silver, and \textsuperscript{\tiny 3}bronze) for accessibility.}
\label{tab:top7performers_percent}
\end{table*}

% As mentioned, in the MAMS dataset, the highest probability pairs performed better than the most common tuples. We hypothesise that this is related to the nature of the dataset, as MAMS is known to contain multiple aspects and sentiments in each example \citep{jiang-etal-2019-challenge}, compared to other datasets.

% Lastly, we performed a backward elimination algorithm \ref{} to find which agents tend to perform across the different datasets and modes...

\paragraph{LLMs Confidence Interpretation}
\label{par:conf}
The \textit{highest confidence list} consistently emerged as the top aggregate technique when using the 72B Qwen; only, in Llama's 70B case in the Restaurant-16 dataset, the \textit{most common list} proved a better strategy. Conversely, while the \textit{lowest confidence list} underperformed significantly with Qwen, it did not score as poorly with Llama. Moreover, when scoring the \textit{most confident agent} amongst the datasets, Qwen's agents two times matched the top performant agent, whereas in Llama's case, it was once. These observations could suggest that the log probabilities provided by Qwen are more informative than those from Llama. The Qwen model is slightly larger, with 72 billion parameters compared to Llama's 70 billion, but it is unclear if this is the sole reason for the observed differences.

One question that we posed to answer was whether the LLM's top performant CoT agent was also the \textit{most confident agent} (i.e. the agent with the sum of highest probability scores). Our results show that although the \textit{most confident agent} would always be amongst the first, it was not always the top performer. In Llama's case, the 70B model only in the MAMS dataset, the top performer was also the most confident, and in Qwen's 72B, two out of four datasets, the top performer was also the most confident. It certainly demonstrates notable consistency, given that there are six different CoT agents that can achieve the highest score. Then, for the 3B LLMs, we got more fuzzy results, with Llama having three out of four and Qwen none out of the four datasets.

\begin{figure}[htbp]
\centering
\includegraphics[width=1\columnwidth]{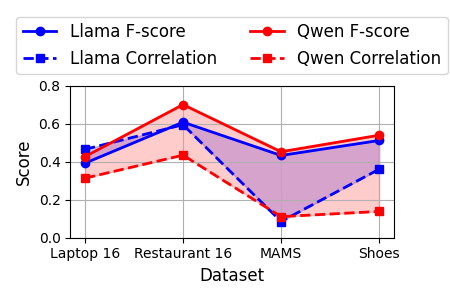}
\caption{Comparison of F-score and Spearman correlation on averaged confidence and F-Score for two LLM models across datasets. Shaded areas represent the difference between the two metrics.}
\label{fig:corr_plot_1}
\end{figure}

To better understand the token-level confidence allocation of each LLM, we analysed the confidence scores of the two larger LLMs. We averaged the probabilities of each pair and then calculated the Spearman correlation \citep{spearman1904} between the scores averaged by each agent and the corresponding F1-score; we performed the same analysis using variance instead of averages. We observed a positive correlation between the averaged probabilities across agents and the F-scores, suggesting that confidence scores indeed reflect a perceived measure of accuracy (see Figure~\ref{fig:corr_plot_1}). Characteristically, Llama exhibited the highest correlation, achieving nearly 60\% correlation on certain datasets. The scores on the MAMS dataset demonstrated a weaker correlation; this could be attributed to the more challenging examples, where sentences contain multiple aspects expressing conflicting emotions \citep{jiang-etal-2019-challenge}.

\begin{figure}[htbp]
    \centering
    \includegraphics[width=1\linewidth]{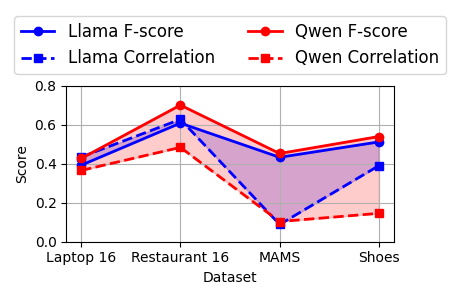}
    \caption{Comparison of F-score and Spearman correlation on confidence variance and F-Score for two LLM models across datasets. Shaded areas represent the difference between the two metrics. The values of the correlation here are swapped with positives. Hence, a correlation in the positive of 0.4 would mean -0.4.}
\label{fig:corr_plot_2}
\end{figure}

Moreover, as shown in Figure~\ref{fig:corr_plot_2}, the variance of probabilities across agents exhibited precisely the same trend as the averaged scores, but inverted. This observation suggests that greater variance among agents is likely associated with a lower F-score.

% One can view the F-scores against the correlations in Appendix~\ref{sec:appendix:correlation_plots}.

% MAMS includes sentences with multiple aspect categories and mixed sentiment polarities. It provides more challenging contexts for ACSA tasks, designed to test the model’s ability to handle sentences where different aspects express conflicting emotions. \citep{jiang-etal-2019-challenge}.

% \paragraph{Prompt Steadiness}

% \paragraph{Decoding Strategies}

% Lastly, on the best found sequence $ Sentiments \rightarrow Categories \rightarrow Aspects $ when firstly extracting via a prompt the relation expressions between aspects and sentiments it scored higher than the best performing element sequence in rest15, however, it was in the middle of performance in rest16 dataset. 
% \paragraph{Order significance}
% I can calculate t-statistic and p-value to know the if what goes first is better

% \paragraph{Multi-Agent CoT Aggregate}

% \paragraph{Attention on Relation Expressions}

% \paragraph{Golden Recipe}
% <best >

% \subsection{Hyper-Parameters \& Generation Issues}

\paragraph{Few Shot CoT}
We evaluated the performance of ChatABSA with 10 few-shot demonstrations against CoT combined with the same 10 few-shot examples. In Appendix~\ref{sec:appendix:few_shot_cot}, we provide further details on how we adapted our CoT prompt to accommodate these few-shot examples.

As expected, when we added 10 examples to ChatABSA, the F-score percentage increased across all datasets. When averaged and compared to Table~\ref{tab:top7performers_percent}, ChatABSA exhibited notable improvements, particularly for the Llama models, with the 3B version increasing by 38\% and the 70B version by 15\%. For the Qwen models, we observed increases of 22\% for the 3B model and 11\% for the 72B model. However, incorporating examples into the CoT resulted in only modest changes: a 4\% increase for Llama 70B, a 3\% decrease for Llama 3B, a 2\% increase for Qwen 72B, and an 8\% increase for Qwen 3B. We observed that the Llama models benefited more substantially from the few-shot demonstrations alone, while the combination of elaborate CoT prompts and examples appeared to ``confuse'' the 3B Llama model. Notably, the Llama 70B model became the only case in which ChatABSA surpassed CoT when averaged across datasets, given the same ten examples as demonstrations.

We hypothesise that the advantage of clear few-shot demonstrations without CoT is that it allows ``the LLMs to focus'' directly on the provided examples and any biases they may introduce without the additional cognitive load imposed by the CoT process, which may divert the model's attention from effectively processing these examples.

% And that's how we fixed the prompt to accommodate the style. We see that it still works the same with CoT added, it improves performance, as shown in the graph below.

\paragraph{CoT Prompt Stability Validation}
We conducted a validation experiment to see the effect of the prompt, comparing the engineered version of the few-shot CoT but without examples (see Appendix~\ref{sec:appendix:few_shot_cot}) to the previously mentioned CoT prompt version (see Appendix~\ref{sec:appendix:cot_prompts}). We ran all models across all datasets with the same system instruction. We observed that, even after changing the prompt, the F1 scores of the 70B+ models remained consistent, differing only from the third decimal point onward. In contrast, for the 3B models, differences appeared from the second decimal point.

% Furthermore, we computed the Pearson correlation coefficient to evaluate whether fluctuations in CoT permutation scores remained consistent between the two prompts for each model and dataset. We found that for the 70B+ models, the Pearson correlation was significant and approximately 90\%, except for the Shoes dataset, which exhibited weaker correlation, likely due to its smaller size, containing only 125 test samples, roughly one-fourth the size of the other datasets. For the 3B models, the Pearson correlation was weaker across all datasets. These results demonstrate that even when the prompt is altered, the CoT element order continues to play a significant role during inference.

\paragraph{Practical Insights}

From Table~\ref{tab:top7performers_percent}, we can see that the \textit{Top performant agent} for each dataset dominates in the higher parameter size LLM models. However, without labelled data, evaluating which CoT order sequence works best for a dataset would be difficult. Following, the \textbf{\textit{Highest probability list} demonstrated favourable results}, indicating that this would be a good choice if no labelled data is available.

We would recommend using various CoT agents and checking the consistency on an unexplored dataset. In Paragraph~\ref{par:conf}, we demonstrated that \textbf{high confidence variance between the agents indicates a more challenging task}. One could leverage this insight to identify challenging reviews within their dataset and examine why the agents' confidence scores deviate among them.

Another valuable metric to include in one’s evaluation repertoire is the extent to which different CoT agents produce \textbf{conflicting categories} (e.g., a category that one agent detects as positive while another agent detects it as negative). As observed, smaller models exhibit this issue more frequently, whereas larger LLMs tend to generate fewer conflicts and consequently achieve higher performance.

Relying on CoT reasoning without input examples offers several advantages over few-shot ICL and fully-supervised approaches. One key benefit is the reduction of biases that can arise from selecting examples. Popular publicly available datasets annotated by multiple crowd-sourced annotators often suffer from annotation bias \citep{geva-etal-2019-modeling,gururangan-etal-2018-annotation,PAULLADA2021100336}. Moreover, within large organisations, if an individual sets the initial examples to address specific cases, these examples may not remain relevant over time. In a dynamic business environment, products, their aspects, and even the workforce evolve, leading to changes in the context and nature of customer feedback. If the initial examples are not regularly updated, they can introduce biases or lead to outdated groupings, ultimately compromising the accuracy and relevance of the analysis.

By contrast, the \textbf{zero-shot approaches do not rely on predefined examples, allowing the model to adapt flexibly to changing contexts}. This approach promotes a more generalised and robust framework for sentiment analysis, reducing the risk of biases and ensuring that the model remains aligned with the current business environment. This adaptability is particularly valuable in scenarios where frequent updates to training data or examples are impractical or resource-intensive.

Another important consideration when applying a zero-shot LLM approach to the ACSA task is \textbf{avoiding overlapping categories in their annotation schema}. It became evident that certain labels tend to be confused; for instance, \textit{contextofuse\#purchase\_context} with \textit{general} in the Shoes dataset, \textit{miscellaneous} with \textit{food} in MAMS, \textit{laptop\#miscellaneous} with \textit{laptop\#general} in Laptop16, and \textit{food\#style options} with \textit{food\#quality} in Restaurant16, among numerous other examples. These overlaps suggest that it is challenging to delineate whether a detected category should fall under an umbrella term (e.g., \textit{general}) or among the more specific labels without access to annotation guidelines or a view of the possible annotators' biases. We recommend that distinct labelling with non-overlapping categories be required when implementing a zero-shot approach.

The advantage of few-shot or fully supervised learning is that it can capture annotators' biases and label the data accordingly. However, we raise the question of how practical this approach is for industry stakeholders who lack the resources to annotate data extensively and require rapid and reproducible results. Therefore, we advocate for the adoption of zero-shot implementations.

\section{Future Direction}
The results demonstrated that the joined CoT agents achieved quite high recall, outperforming any individual CoT agent. This demonstrates that there is potential for improvement in the use of agents. One possible direction for future research could be to iteratively guide the inference process until a satisfactory summed log probability is reached. Moreover, one can try combining other CoT agents to evaluate possible improvements. Or utilising other aggregating techniques to filter the pairs.

Research has raised concerns regarding the reliability of token-level uncertainty, as token conditional distributions may sometimes be misleading. This issue arises particularly in cases where an initial token is incorrect, yet all subsequent tokens appear highly probable given that initial token \citep{shorinwa2024surveyuncertaintyquantificationlarge}. Investigating more robust methods for estimating token-level confidence scores could improve the downstream ACSA performance or other tasks as well.

% - say these things are more wide than ABSA

% Although this approach is effective in general, the conditional distribution of the tokens can be misleading in certain scenarios, especially when an initial token is incorrect but all the succeeding tokens are highly probable given the initial token. 

\section{Conclusion}
 In our experiments, we evaluated two larger (70B+) LLMs, Qwen and Llama, along with their smaller 3B parameter-size counterparts. We observed that the optimal CoT order for the ACSA task is rather dataset-dependent. Additionally, we found that employing a multi-hop CoT approach does not yield better results compared to our enumerated CoT version. Furthermore, confidence scores derived from token-level uncertainty proved somewhat informative, appearing to be a more effective strategy than relying solely on majority voting among agents. Moreover, we discovered that using the variance in confidence between agents can be a good indicator to view challenging examples. Intuitively, larger models tend to produce more reliable confidence estimates. Given the high recall achieved by our combined CoT aggregation method, we suggest there remains considerable room for improvement, potentially through exploring alternative aggregation techniques or developing more accurate token-level uncertainty estimation methods.

% - order is rather data driven

% More and better split datasets.
% - prob > voting
% - multi-hop not better
% - Larger models have better uncertainty representation

\section{Limitations}
% One limitation is the reliance on multiple paraphrases of prompting techniques. It is known that the output of an LLM may vary depending on the prompt used. However, our empirical studies have indicated that these variations are consistent enough not to significantly impact the results. Similarly, we did not test different system instruction formats.

One limitation is the primacy bias or order bias effect \citep{wang-etal-2023-primacy}. LLMs' generation may get biased towards an earlier exposed option when asked, rather than a later one. This bias is particularly prevalent in category selection; when we ask the LLM which categories are mentioned in the text, it may have a tendency to select options presented at the beginning, ideally, we would need to change the order of the categories, but this would be computationally expensive to investigate.

% Furthermore, we do not know the extent to which each LLM has been exposed to each dataset during its training, and how this exposure has impacted their F-score performance. Despite this, we have taken these factors into consideration and formatted the results accordingly, avoiding substantial claims based on these potential biases.

% Different types of prompting.

% Dataset availability.

% Ask LLM whether they now the datasets and the task.

% Maybe if there is more word generation, it could have better results.

% Do hyper-parameter tuning.

% Some would not generate because of some harasment issues, especially with GEMINI.

% Category some are present in the ICL, and some are not.

% Prompts, other wordings such as conjuction words vs relationship epxressions.
% Didn't test different system instruction formats.
% Primacy bias, like in categories.

% add definitions on the categories

\section{Ethical Statement}
LLMs, due to their training on extensive amounts of internet data, may inadvertently perpetuate and amplify biases present in the source material. This could result in biased or unfair outcomes in ABSA tasks, particularly in sensitive domains such as social media monitoring.

Our research contributes to the enhancement of sentiment analysis techniques, which can be beneficial in various fields such as marketing, customer service, and social sciences. By providing more accurate sentiment analysis, organisations and researchers can make better-informed decisions based on public and customer opinions.

However, there is a risk that enhanced ABSA techniques could be misused for unethical purposes, such as manipulating public opinion or spreading misinformation. We advocate for responsible use and adherence to ethical standards in deploying such technologies.

% \section*{Acknowledgements}
% TBD
% AT for funding

\bibliography{output}

% \clearpage
\appendix

\section{Number of Pair Estimation - Alpha}
\label{sec:appendix:alpha}
% For the aggregation techniques \textit{highest probability pairs} and \textit{clustered pairs} in Section~\ref{sec:aggregation_techniques}, we came up with some techniques for calculating the number of pairs to collect from the CoT agents for a specific instance, we call this variable $alpha$. Following we enumerate the techniques:
For the aggregation techniques \textit{highest probability pairs} and \textit{clustered pairs} described in Section~\ref{sec:aggregation_techniques}, we developed several methods to determine the number of pairs to collect from the CoT agents for a given instance; we denote this variable as $\alpha$. These methods are enumerated as follows:

\paragraph{Float Variable}
Given a float variable, it would act as a variable of bias between the number of pairs produced by the CoT agents and the global average of the number of pairs produced in the whole dataset. In the below equation, the $\mu$ represents the median of all the pairs on all the instances produced by the CoT agents, and the left part of the equation represents the average number of pairs for the current instance.

\[
\text{round} \left( \alpha \cdot \frac{1}{N} \sum_{i=1}^{N} n_{\text{pairs}, i} + (1 - \alpha) \cdot \mu \right)
\]

In our study, we experimented with the values $0.9$ and $1$.

\paragraph{String Variable}
% The other process would involve acquiring the average log prob of each agent and mapping them against the number of pairs they produced. For instance, if the first agent had two pairs then we would map 2 with the average of the log-probs like 0.9. Then, once we have a number of pairs with their corresponding averages, say we have four 2 and two 3s, we would either do \textit{mean} or \textit{max} on the the 2's and 3's averaged probs. Finally, we would select the one with the highest score. Thus, that is where the \textit{mean} and \textit{max} correspond when configuring the $alpha$ score.

The other process involves first computing the average (list) probability for each agent and then associating this average with the number of pairs the agent produces. For instance, if an agent produces two pairs and has an average probability of 0.9, we record the mapping (2, 0.9). Once we collect these (number of pairs: average log probability) mappings for all agents, we subsequently apply either the \textit{mean} or \textit{max} operation to the mapped probabilities. Finally, we select the number of pairs with the highest value. This illustrates how the \textit{mean} and \textit{max} variables operate when configuring the $alpha$ score.

\section{Dataset Statistics}
\label{sec:appendix:dataset_statistics}
In Table~\ref{tab:datasets} we show the dataset statistics.
% Making it H or table* does not work, perhaps I can add below 2 column width more tables
\begin{table*}[h]
\centering
\begin{tabular}{lllll}
\hline
Datasets        & Laptop 16 & Restaurant 16 & MAMS & Shoes \\ \hline
N. train samples     &  2468  & 1954 & 3149 & 906   \\
N. val samples       &  n/a   & n/a & 400 & 116    \\
N. test samples      &  579   & 571 & 400 & 125   \\
N. categories        &  67    & 12  & 8   & 21  \\ \hline
\end{tabular}
\caption{Dataset statistics for the four datasets employed in our study. The number of instances comes after pre-processing, omitting any examples with conflicting labels. Laptop16 and Restaurant16 do not have a validation dataset.}
\label{tab:datasets}
\end{table*}

\section{CoT Agent Performance}
\label{sec:appendix:cot_agent_performance}

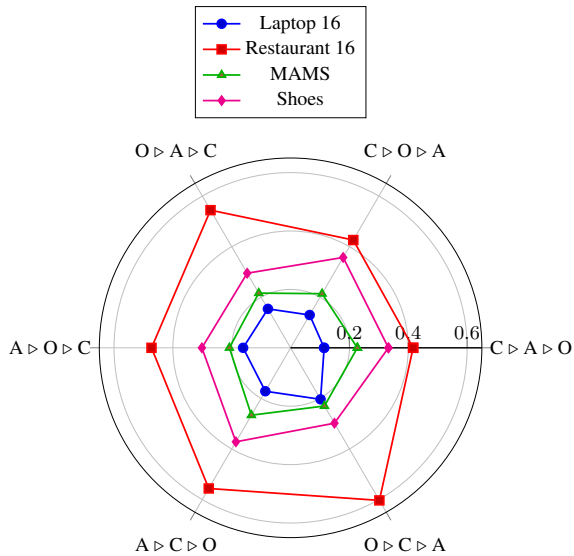
\begin{figure}[htbp]
\centering
\resizebox{\columnwidth}{!}{%

\begin{tikzpicture}
\begin{polaraxis}[
    width=\columnwidth,
    height=\columnwidth,
    xtick={0,60,...,300},
    xticklabels={
        C $\triangleright$ A $\triangleright$ O,
        C $\triangleright$ O $\triangleright$ A,
        O $\triangleright$ A $\triangleright$ C,
        A $\triangleright$ O $\triangleright$ C,
        A $\triangleright$ C $\triangleright$ O,
        O $\triangleright$ C $\triangleright$ A
    },
    ytick={0.2,0.4,0.6},
    ymin=0, ymax=0.65,
    legend style={at={(0.25,1.4)},anchor=north west,font=\small},
    grid=both,
    minor grid style={gray!25},
    major grid style={gray!50},
    tick label style={font=\small},
    label style={font=\small},
]

% Laptop 16
\addplot+[blue,mark=*,thick] coordinates {
    (0,0.1140) % C>A>O
    (60,0.1297) % C>O>A
    (120,0.1540) % O>A>C
    (180,0.1614) % A>O>C
    (240,0.1717) % A>C>O
    (300,0.2037) % O>C>A
    (360,0.1140) % close loop
};
\addlegendentry{Laptop 16}

% Restaurant 16
\addplot+[red,mark=square*,thick] coordinates {
    (0,0.4165) % C>A>O
    (60,0.4262) % C>O>A
    (120,0.5443) % O>A>C
    (180,0.4727) % A>O>C
    (240,0.5562) % A>C>O
    (300,0.6034) % O>C>A
    (360,0.4165) % close loop
};
\addlegendentry{Restaurant 16}

% MAMS
\addplot+[green!70!black,mark=triangle*,thick] coordinates {
    (0,0.2269) % C>A>O
    (60,0.2140) % C>O>A
    (120,0.2163) % O>A>C
    (180,0.2080) % A>O>C
    (240,0.2664) % A>C>O
    (300,0.2301) % O>C>A
    (360,0.2269) % close loop
};
\addlegendentry{MAMS}

% Shoes
\addplot+[magenta,mark=diamond*,thick] coordinates {
    (0,0.3325) % C>A>O
    (60,0.3573) % C>O>A
    (120,0.2950) % O>A>C
    (180,0.3014) % A>O>C
    (240,0.3717) % A>C>O
    (300,0.2981) % O>C>A
    (360,0.3325) % close loop
};
\addlegendentry{Shoes}

\end{polaraxis}
\end{tikzpicture}
}
\caption{Radar plot comparing F-scores across different datasets and element orders for the Qwen 3B model.}
\label{fig:radar_fscore}
\end{figure}

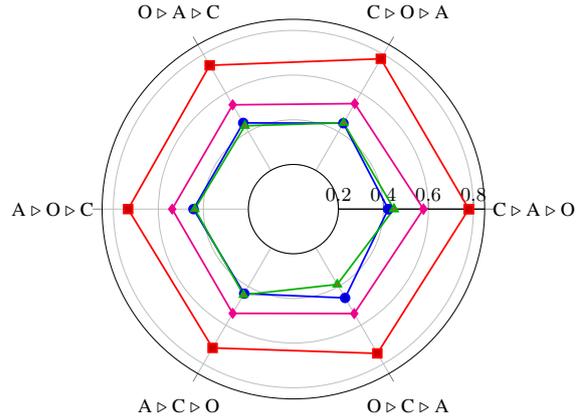
\begin{figure}[htbp]
\centering
\resizebox{\columnwidth}{!}{%

\begin{tikzpicture}
\begin{polaraxis}[
    width=\columnwidth,
    height=\columnwidth,
    xtick={0,60,...,300},
    xticklabels={
        C $\triangleright$ A $\triangleright$ O,
        C $\triangleright$ O $\triangleright$ A,
        O $\triangleright$ A $\triangleright$ C,
        A $\triangleright$ O $\triangleright$ C,
        A $\triangleright$ C $\triangleright$ O,
        O $\triangleright$ C $\triangleright$ A
    },
    ytick={0.2,0.4,0.6,0.8},
    ymin=0.2, ymax=0.85,
    legend style={at={(1.05,0.95)},anchor=north west,font=\small},
    grid=both,
    minor grid style={gray!25},
    major grid style={gray!50},
    tick label style={font=\small},
    label style={font=\small},
]

% Laptop 16
\addplot+[blue,mark=*,thick] coordinates {
    (0,0.4216) % C>A>O
    (60,0.4440) % C>O>A
    (120,0.4457) % O>A>C
    (180,0.4440) % A>O>C
    (240,0.4371) % A>C>O
    (300,0.4593) % O>C>A
    (360,0.4216) % close loop
};
% \addlegendentry{Laptop 16}

% Restaurant 16
\addplot+[red,mark=square*,thick] coordinates {
    (0,0.7796) % C>A>O
    (60,0.7773) % C>O>A
    (120,0.7435) % O>A>C
    (180,0.7345) % A>O>C
    (240,0.7179) % A>C>O
    (300,0.7462) % O>C>A
    (360,0.7796) % close loop
};
% \addlegendentry{Restaurant 16}

% MAMS
\addplot+[green!70!black,mark=triangle*,thick] coordinates {
    (0,0.4471) % C>A>O
    (60,0.4459) % C>O>A
    (120,0.4307) % O>A>C
    (180,0.4381) % A>O>C
    (240,0.4433) % A>C>O
    (300,0.3896) % O>C>A
    (360,0.4471) % close loop
};
% \addlegendentry{MAMS}

% Shoes
\addplot+[magenta,mark=diamond*,thick] coordinates {
    (0,0.5779) % C>A>O
    (60,0.5455) % C>O>A
    (120,0.5392) % O>A>C
    (180,0.5387) % A>O>C
    (240,0.5392) % A>C>O
    (300,0.5403) % O>C>A
    (360,0.5779) % close loop
};
% \addlegendentry{Shoes}

\end{polaraxis}
\end{tikzpicture}
}
\caption{Radar plot comparing F-scores across different datasets and element orders for the Qwen 72B model.}
\label{fig:radar_fscore_second}
\end{figure}

\begin{figure}[htbp]
\centering
\resizebox{\columnwidth}{!}{%

\begin{tikzpicture}
\begin{polaraxis}[
    width=\columnwidth,
    height=\columnwidth,
    xtick={0,60,...,300},
    xticklabels={
        C $\triangleright$ A $\triangleright$ O,
        C $\triangleright$ O $\triangleright$ A,
        O $\triangleright$ A $\triangleright$ C,
        A $\triangleright$ O $\triangleright$ C,
        A $\triangleright$ C $\triangleright$ O,
        O $\triangleright$ C $\triangleright$ A
    },
    ytick={0.2,0.3,0.4,0.5},
    ymin=0.1, ymax=0.5,
    legend style={at={(1.05,0.95)},anchor=north west,font=\small},
    grid=both,
    minor grid style={gray!25},
    major grid style={gray!50},
    tick label style={font=\small},
    label style={font=\small},
]

% Laptop 16
\addplot+[blue,mark=*,thick] coordinates {
    (0,0.1028) % C>A>O
    (60,0.1559) % C>O>A
    (120,0.1912) % O>A>C
    (180,0.1764) % A>O>C
    (240,0.1176) % A>C>O
    (300,0.1605) % O>C>A
    (360,0.1028) % close loop
};
% \addlegendentry{Laptop 16}

% Restaurant 16
\addplot+[red,mark=square*,thick] coordinates {
    (0,0.2530) % C>A>O
    (60,0.2556) % C>O>A
    (120,0.4318) % O>A>C
    (180,0.3244) % A>O>C
    (240,0.2945) % A>C>O
    (300,0.3377) % O>C>A
    (360,0.2530) % close loop
};
% \addlegendentry{Restaurant 16}

% MAMS
\addplot+[green!70!black,mark=triangle*,thick] coordinates {
    (0,0.2103) % C>A>O
    (60,0.2006) % C>O>A
    (120,0.2097) % O>A>C
    (180,0.1783) % A>O>C
    (240,0.1956) % A>C>O
    (300,0.1861) % O>C>A
    (360,0.2103) % close loop
};
% \addlegendentry{MAMS}

% Shoes
\addplot+[magenta,mark=diamond*,thick] coordinates {
    (0,0.1454) % C>A>O
    (60,0.1813) % C>O>A
    (120,0.2154) % O>A>C
    (180,0.2181) % A>O>C
    (240,0.2428) % A>C>O
    (300,0.2167) % O>C>A
    (360,0.1454) % close loop
};
% \addlegendentry{Shoes}

\end{polaraxis}
\end{tikzpicture}
}
\caption{Radar plot comparing F-scores across different datasets and element orders for the Llama 3B model.}
\label{fig:radar_fscore_third}
\end{figure}
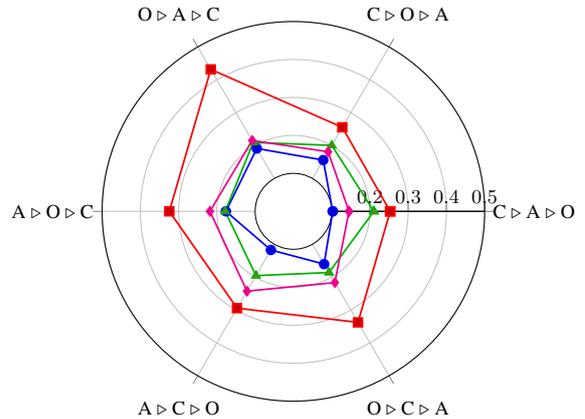

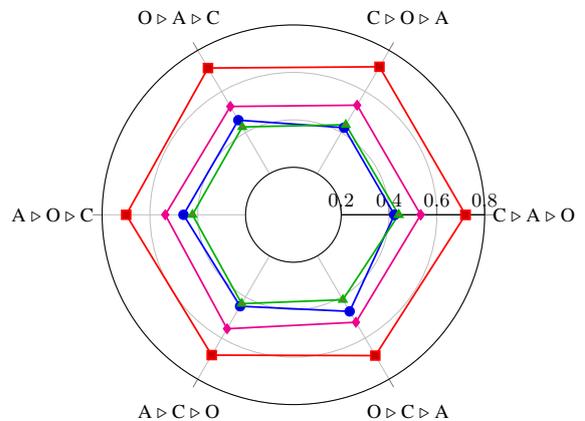
\begin{figure}[htbp]
\centering
\resizebox{\columnwidth}{!}{%
\begin{tikzpicture}
\begin{polaraxis}[
    width=\columnwidth,
    height=\columnwidth,
    xtick={0,60,...,300},
    xticklabels={
        C $\triangleright$ A $\triangleright$ O,
        C $\triangleright$ O $\triangleright$ A,
        O $\triangleright$ A $\triangleright$ C,
        A $\triangleright$ O $\triangleright$ C,
        A $\triangleright$ C $\triangleright$ O,
        O $\triangleright$ C $\triangleright$ A
    },
    ytick={0.2,0.4,0.6,0.8},
    ymin=0.2, ymax=0.8,
    legend style={at={(1.05,0.95)},anchor=north west,font=\small},
    grid=both,
    minor grid style={gray!25},
    major grid style={gray!50},
    tick label style={font=\small},
    label style={font=\small},
]

% Laptop 16
\addplot+[blue,mark=*,thick] coordinates {
    (0,0.4227) % C>A>O
    (60,0.4238) % C>O>A
    (120,0.4605) % O>A>C
    (180,0.4598) % A>O>C
    (240,0.4451) % A>C>O
    (300,0.4702) % O>C>A
    (360,0.4227) % close loop
};
% \addlegendentry{Laptop 16}

% Restaurant 16
\addplot+[red,mark=square*,thick] coordinates {
    (0,0.7198) % C>A>O
    (60,0.7202) % C>O>A
    (120,0.7141) % O>A>C
    (180,0.6994) % A>O>C
    (240,0.6830) % A>C>O
    (300,0.6854) % O>C>A
    (360,0.7198) % close loop
};
% \addlegendentry{Restaurant 16}

% MAMS
\addplot+[green!70!black,mark=triangle*,thick] coordinates {
    (0,0.4405) % C>A>O
    (60,0.4383) % C>O>A
    (120,0.4274) % O>A>C
    (180,0.4222) % A>O>C
    (240,0.4325) % A>C>O
    (300,0.4136) % O>C>A
    (360,0.4405) % close loop
};
% \addlegendentry{MAMS}

% Shoes
\addplot+[magenta,mark=diamond*,thick] coordinates {
    (0,0.5326) % C>A>O
    (60,0.5332) % C>O>A
    (120,0.5271) % O>A>C
    (180,0.5359) % A>O>C
    (240,0.5547) % A>C>O
    (300,0.5230) % O>C>A
    (360,0.5326) % close loop
};
% \addlegendentry{Shoes}

\end{polaraxis}
\end{tikzpicture}
}
\caption{Radar plot comparing F-scores across different datasets and element orders for the Llama 70B model.}
\label{fig:radar_fscore_fourth}
\end{figure}

In Figures ~\ref{fig:radar_fscore}, \ref{fig:radar_fscore_second}, \ref{fig:radar_fscore_third}, \ref{fig:radar_fscore_fourth} we show the various CoT F1 scores per dataset, as shown in the legend. Each figure is a different model. The letters A, C and O stand for Aspects, Category and Opinion, respectively and the arrows in between denote their order in the CoT prompt.

\section{Joined CoT Agents}
\label{sec:appendix:joined}
Below, in Figure \ref{fig:labels_confl}, we report the number of conflicting pairs for each dataset, after combining the pairs produced by the CoT agents, also referred to as the \textit{Joined CoT Agent}.

% \begin{figure*}[h]
%     \centering
%     \includegraphics[width=0.9\linewidth]{figures/n_labels_new.png}
%     \caption{The average number of pairs generated for each model for each dataset. These refer to the naive model aggregate, the pairs are summed. In dark grey is the average number of the ground truth for each dataset.}
%     \label{fig:n_labels}
% \end{figure*}

\begin{figure*}
    \centering
    \includegraphics[width=0.9\linewidth]{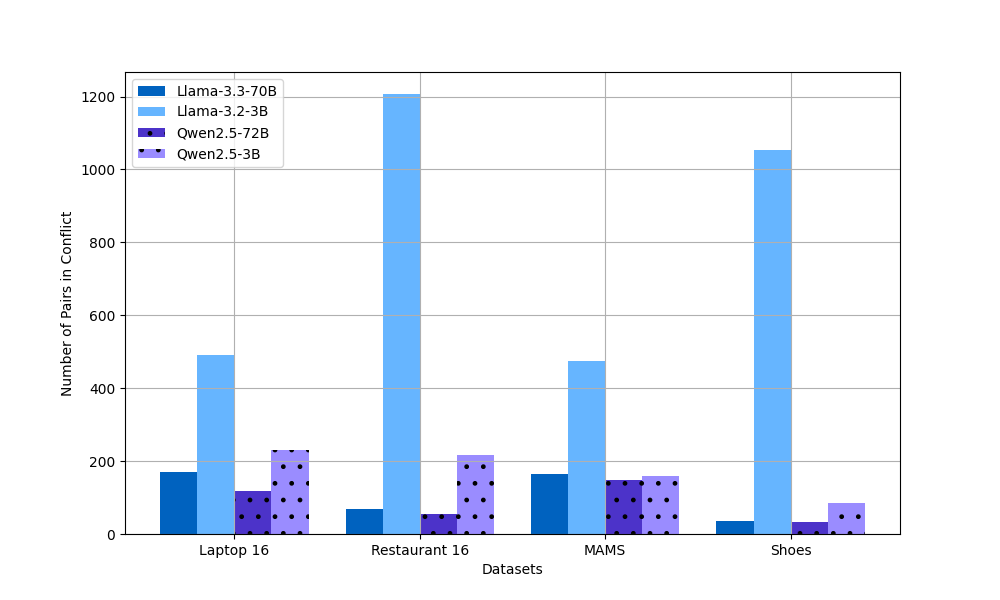}
    \caption{Number of conflicting pairs for each dataset per model for the Joined CoT Agent.}
    \label{fig:labels_confl}
\end{figure*}

\clearpage
% \section{LM versions}
% \label{sec:appendix:versions}
% {\small
% \begin{lstlisting}[language=, caption={LM versions used in our experiments:}, label=lst:llm_versions]
%    - Qwen/Qwen2.5-72B-Instruct
%    - Qwen/Qwen2.5-3B-Instruct
%    - meta-llama/Llama-3.3-70B-Instruct
%    - meta-llama/Llama-3.2-3B-Instruct
% \end{lstlisting}
% }

\clearpage

\section{CoT Prompt}
\label{sec:appendix:cot_prompts}
The prompts of our CoT agents are compositional in the sense that each element has its own template, whether it is the first or a subsequent element in the chain. As previously discussed, we have three element permutations consisting of the aspect term, aspect category, and sentiment polarity. Thus, we have $3 \times 2 = 6$ templates, which we chain according to the given order to form the CoT agent's prompt. Below, we provide an example of the Aspect $\rightarrow$ Opinion $\rightarrow$ Category ordering. As shown, the instructions for each element are numbered and presented sequentially. The prompt for the first element differs in that it includes the review text and begins with the phrase ``Given the following text''. After these instructions, we prompt the LLM to provide answers by enumerating the elements and including blanks for the LLM to fill in. Finally, we instruct the model to produce the ACSA tuples formatted as a Python list of tuples. 

We obtained more stable results by using different types of brackets when specifying the output, i.e., square brackets and parentheses, rather than nested square brackets. The same logic applies to quotation marks: to avoid obfuscating the instructions, we use double quotes around the example demonstration list of tuples and enclose the strings within single quotes.

Additionally, we first present the system instruction used in our study, adapted from \citep{10499502}, which employs a role-playing format and constrains the LLM to generate the most probable answers while controlling their verbosity.

The fonts in the examples are for illustration purposes only.

\resizebox{\columnwidth}{!}{%
\begin{tcolorbox}[colframe=black, colback=gray!10, width=3.5in, title=System Instruction]
\begin{tikzpicture}
\node[draw, rounded corners, fill=orange!10, text width=1\columnwidth, align=left] at (0,0) (system) {
You are a Natural Language Processing assistant, expert in Aspect-Based Sentiment Analysis. I want you to force yourself to pick words that you are being asked and only them, without explanations or reasoning. If you are unsure, put the most probable. Now follow the following steps:
};
\end{tikzpicture}
\end{tcolorbox}
}
\vspace{1em}

\begin{adjustbox}{width=\columnwidth}

\begin{tcolorbox}[colframe=black, colback=gray!10, width=3.5in, title=CoT Prompt]
\begin{tikzpicture}
\node[draw, rounded corners, fill=blue!10,, text width=1\columnwidth, align=left] at (0,0) (prompt) {
1. Given the following text, list all word sequences that denote an aspect term of the restaurant domain:\\[0.5em]
\textit{``We went again and sat at the bar this time, I had 5 pints of guinness and not one buy-back, I ordered a basket of onion rings and there were about 5 in the basket, the rest was filled with crumbs, the chili was not even edible.''}\\[0.5em]
2. List all word sequences that denote or link to an opinion from the aspects detected.\\[0.5em]
3. List the categories from the opinions detected. The list of possible categories is: \verb|[menu, service, price, ambience, |

\verb|place, staff, miscellaneous, food]|.\\[0.5em]
\textbf{-----------------------------------}\\[0.5em]
1. Aspects:\\[0.5em]
2. Opinions:\\[0.5em]
3. Categories:\\[0.5em]
Lastly, please provide one Python-type list of tuples such as\\[0.5em]
\verb|''[('example\_category_1', 'positive'),|
\verb|('example_category_2', 'negative'), ...]''|\\[0.5em]
where the categories are provided above and the sentiment is either \textit{positive}, \textit{neutral}, or \textit{negative}, based on the extracted opinions.
};
\end{tikzpicture}
\end{tcolorbox}
\end{adjustbox}

\clearpage
\section{Multi-Hop CoT Prompt}
\label{sec:appendix:multi_hop}
To compare with previous CoT research in the ABSA domain, we adapted the multi-hop approach of \citet{fei-etal-2023-reasoning,10499502} to our ACSA context. Unlike \citet{fei-etal-2023-reasoning}, who first provide the aspect term to search, then ask the LLM to identify it from the text, and subsequently use this answer to extract the opinion from the text, which they finally feed into the last prompt to determine the sentiment polarity, we do not assume prior knowledge of the targets. Instead, we follow the same sequential approach as illustrated in our examples in Appendix~\ref{sec:appendix:cot_prompts}, but in a multi-hop manner. After each step, we pose a question to the LLM, and the retrieved answer is added to the message history using the chat template. Below, we demonstrate an example using the 3B Llama model as the assistant, which ultimately provides the ACSA tuples as a Python-style list of tuples for post-processing. Consequently, we make four calls to the LLM for one prediction. Additionally, we would use the same system instruction as in Appendix~\ref{sec:appendix:cot_prompts}.

\begin{adjustbox}{width=\columnwidth}
\begin{tcolorbox}[colframe=black, colback=gray!10, width=3.5in, title=CoT Multi-Hop Thread]
\begin{tikzpicture}[node distance=0.4cm]

% First node (User, left-aligned, default position)
\node[draw, rounded corners, fill=blue!10, text width=0.95\columnwidth, align=left] (user1) {
\textbf{User:}\\[0.3em]
Given the following text, list all word sequences that denote an aspect term of the restaurant domain:\\[0.3em]
\textit{``We went again and sat at the bar this time, I had 5 pints of guinness and not one buy-back, I ordered a basket of onion rings and there were about 5 in the basket, the rest was filled with crumbs, the chili was not even edible.''}
};

% Assistant (right-aligned, shifted right)
\node[draw, rounded corners, fill=green!10, text width=0.5\columnwidth, align=left, below=of user1, xshift=2cm] (assistant1) {
\textbf{Assistant:}\\[0.3em]
bar, onion, rings, chili
};

% User (left-aligned, shifted left)
\node[draw, rounded corners, fill=blue!10, text width=0.95\columnwidth, align=left, below=of assistant1, xshift=-2cm] (user2) {
\textbf{User:}\\[0.3em]
List all word sequences that denote or link to an opinion from the aspects detected.
};

% Assistant (right-aligned, shifted right)
\node[draw, rounded corners, fill=green!10, text width=0.5\columnwidth, align=left, below=of user2, xshift=2cm] (assistant2) {
\textbf{Assistant:}\\[0.3em]
not, even, edible
};

% User (left-aligned, shifted left)
\node[draw, rounded corners, fill=blue!10, text width=0.95\columnwidth, align=left, below=of assistant2, xshift=-2cm] (user3) {
\textbf{User:}\\[0.3em]
List the categories from the opinions detected. The list of possible categories is:\\
\verb|['menu', 'service', 'price',|\\
\verb|'ambience', 'place', 'staff',|\\
\verb|'miscellaneous', 'food']|
};

% Assistant (right-aligned, shifted right)
\node[draw, rounded corners, fill=green!10, text width=0.5\columnwidth, align=left, below=of user3, xshift=2cm] (assistant3) {
\textbf{Assistant:}\\[0.3em]
\verb|['menu', 'food']|
};

% User (left-aligned, shifted left)
\node[draw, rounded corners, fill=blue!10, text width=1\columnwidth, align=left, below=of assistant3, xshift=-2cm] (user4) {
\textbf{User:}\\[0.3em]
Lastly, please provide one Python type list of tuples such as:\\
\verb|[('example_category_1', 'positive'),|\\
\verb| ('example_category_2', 'negative'), ...]|\\
that you identified. Where the categories are provided above and the sentiment is either 'positive', 'neutral' or 'negative', based on the extracted opinions.
};

% Assistant (right-aligned, shifted right)
\node[draw, rounded corners, fill=green!10, text width=0.55\columnwidth, align=left, below=of user4, xshift=1.8cm] (assistant4) {
\textbf{Assistant:}\\[0.3em]
\verb|[('menu', 'negative'),|\\
\verb| ('food', 'negative')]|
};

\end{tikzpicture}
\end{tcolorbox}
\end{adjustbox}

\section{Few Shot CoT}
\label{sec:appendix:few_shot_cot}

The prompt used to construct the few-shot CoT prompt, incorporating ten examples from the training dataset, required careful engineering. Within the CoT, the elements—aspect terms, opinions, and categories—are enumerated in varying orders. However, since the datasets employed in this study are ACSA-type, the ground-truth annotations contain only categories and polarities without explicit annotations for aspect terms and opinions. Consequently, the prompt was designed to explicitly illustrate the internal CoT reasoning steps that the LLM should follow between clearly defined BEGIN COT and END COT markers, subsequently instructing the model to output the final result explicitly as a Python list. Given the complexity and detailed nature of this prompt, additional precision was required, including explicit reminders to output the Python list after completing the CoT reasoning steps; this explicit instruction was deemed particularly necessary for the 3B parameter models. Below is a prompt containing two examples.

\begin{adjustbox}{width=\columnwidth}

\begin{tcolorbox}[colframe=black, colback=gray!10, width=3.5in, title=Few Shot CoT Prompt (1/2)]
\begin{tikzpicture}
\node[draw, rounded corners, fill=blue!10, text width=1\columnwidth, align=left] at (0,0)(prompt){
{\small
Given the following text, list all word sequences that denote an aspect term of the restaurant domain:\\[0.5em]
\textit{"Worst Service I Ever Had"}\\[0.5em]

List all word sequences that denote or link to an opinion from the aspects detected.\\[0.5em]

List the categories from the opinions detected. The list of possible categories is:\\[0.5em]

\verb|['FOOD#QUALITY', 'AMBIENCE#GENERAL',|\
\verb| 'SERVICE#GENERAL', 'RESTAURANT#PRICES',|\
\verb|'DRINKS#QUALITY', 'FOOD#PRICES',|\
\verb|'RESTAURANT#MISCELLANEOUS', 'LOCATION#GENERAL',|\
\verb|'DRINKS#STYLE_OPTIONS', 'DRINKS#PRICES',|\
\verb|'FOOD#STYLE_OPTIONS', 'RESTAURANT#GENERAL']|.\\[0.5em]

• The reasoning must appear only between BEGIN COT and END COT.\\[0.5em]

-------\\[0.5em]
BEGIN COT\\[0.5em]

Aspects: ...\\[0.5em]
Opinions: ...\\[0.5em]
Categories: ...\\[0.5em]
END COT\\[0.5em]
-------\\[0.5em]

• Outside those markers print one PYTHON LIST of tuples, exactly like\\[0.5em]
\verb|[('example_category_1', 'negative'),|\
\verb|('example_category_2', 'positive')]| \
that you identify in your three step COT reasoning.\\[0.5em]

• Where the categories derive from step 3 Categories in COT and each associated category's sentiment is either 'positive', 'neutral' or 'negative', based on step 2 the extracted Opinions.\\[0.5em]

}
}; 
\end{tikzpicture} 
\end{tcolorbox}
\end{adjustbox}

\begin{adjustbox}{width=\columnwidth}
\begin{tcolorbox}[colframe=black, colback=gray!10, width=3.5in, title=Few Shot CoT Prompt (2/2)]
\begin{tikzpicture}
\node[draw, rounded corners, fill=blue!10, text width=1\columnwidth, align=left] at (0,0)(prompt){
{\small

----------------------------\\[0.5em]
EXAMPLES\\[0.5em]
----------------------------\\[0.5em]

Example 1\\[0.5em]
Review: \textit{"Service was wonderful;"}\\[0.5em]
-------\\[0.5em]
BEGIN COT\\[0.5em]

Aspects: ...\\[0.5em]
Opinions: ...\\[0.5em]
Categories: ...\\[0.5em]
END COT\\[0.5em]
-------\\[0.5em]
PYTHON LIST: \verb|[('SERVICE#GENERAL', 'positive')]|\\[0.5em]

Example 2\\[0.5em]
Review: \textit{"My mom originally introduced me to this place, but even she (being Indian) feels the food can be somewhat over the top spicy and far too oily."}\\[0.5em]
-------\\[0.5em]
BEGIN COT\\[0.5em]

Aspects: ...\\[0.5em]
Opinions: ...\\[0.5em]
Categories: ...\\[0.5em]
END COT\\[0.5em]
-------\\[0.5em]
PYTHON LIST: \verb|[('FOOD#QUALITY', 'negative')]|\\[0.5em]

----------------------------\\[0.5em]
NOW SOLVE THE NEW REVIEW\\[0.5em]
----------------------------\\[0.5em]

• Please complete the below CoT and end it with END COT.\\[0.5em]

Given the following text, list all word sequences that denote an aspect term of the restaurant domain:\\[0.5em]
\textit{"Worst Service I Ever Had"}\\[0.5em]

List all word sequences that denote or link to an opinion from the aspects detected.\\[0.5em]

List the categories from the opinions detected. The list of possible categories is:\\[0.5em]
\verb|['FOOD#QUALITY', 'AMBIENCE#GENERAL',|\
\verb| 'SERVICE#GENERAL', 'RESTAURANT#PRICES',|\
\verb|'DRINKS#QUALITY', 'FOOD#PRICES',|\
\verb|'RESTAURANT#MISCELLANEOUS', 'LOCATION#GENERAL',|\
\verb|'DRINKS#STYLE_OPTIONS', 'DRINKS#PRICES',|\
\verb|'FOOD#STYLE_OPTIONS', 'RESTAURANT#GENERAL']|.\\[0.5em]

• After completing the three step COT reasoning and closing it with END COT, print the PYTHON LIST of tuples, exactly like \
\verb|[('example_category_1', 'negative'),| \
\verb|('example_category_2', 'positive')]| \
that you identify in your three step COT reasoning.\\[0.5em]

• Where the categories derive from step 3 Categories in COT and each associated category's sentiment is either 'positive', 'neutral' or 'negative', based on step 2 the extracted Opinions.\\[0.5em]

• DO NOT FORGET the PYTHON LIST.\\[0.5em]

-------\\[0.5em]
BEGIN COT\\[0.5em]

Aspects: ... 
}
}; 
\end{tikzpicture} 
\end{tcolorbox}
\end{adjustbox}

\section{CoT versus Multi-Hop CoT}
\label{sec:appendix:cot_vs_multi_hop}
We conducted experiments using the 3B LLMs to determine whether enumerating tasks within a single CoT prompt (see example in Appendix~\ref{sec:appendix:cot_prompts}) or using four separate calls via the multi-hop CoT approach (see example in Appendix~\ref{sec:appendix:multi_hop}) yields higher prediction scores. Table~\ref{tab:cot_vs_multi_hop} presents the best-performing CoT agent for each approach, organised by LLM and dataset. Recall that we have six different agents, each varying the order of elements. As shown in the table, one notable characteristic of the multi-hop CoT approach is that Opinions always precede Categories and Aspect terms.

% \usepackage{graphicx}
\begin{table*}[h]
\centering

\resizebox{\textwidth}{!}{%
\begin{tabular}{lllllllll}
\multicolumn{1}{c}{Datasets}      & \multicolumn{2}{c}{Laptop 16}                                                               & \multicolumn{2}{c}{Restaurant16}                                                            & \multicolumn{2}{c}{MAMS}                                                                    & \multicolumn{2}{c}{Shoes}                                                                   \\ \hline
\multicolumn{1}{c}{3B LLM models} & \multicolumn{1}{c}{Qwen}                     & \multicolumn{1}{c}{Llama}                    & \multicolumn{1}{c}{Qwen}                     & \multicolumn{1}{c}{Llama}                    & \multicolumn{1}{c}{Qwen}                     & \multicolumn{1}{c}{Llama}                    & \multicolumn{1}{c}{Qwen}                     & \multicolumn{1}{c}{Llama}                    \\ \hline
Enumerated CoT                    & O \textrangle C \textrangle A: \textbf{20.3} & O \textrangle A \textrangle C: \textbf{19.1} & O \textrangle C \textrangle A: \textbf{60.0} & O \textrangle A \textrangle C: 43.1          & A \textrangle C \textrangle O: 26.6          & C \textrangle A \textrangle O: 21.0          & A \textrangle C \textrangle O: \textbf{37.1} & A \textrangle C \textrangle O: 24.2          \\
Multi-hop CoT                     & O \textrangle C \textrangle A: 15.9          & O \textrangle C \textrangle A: 17.1          & O \textrangle C \textrangle A: 56.5          & O \textrangle C \textrangle A: \textbf{52.5} & O \textrangle C \textrangle A: \textbf{28.1} & O \textrangle C \textrangle A: \textbf{29.1} & O \textrangle C \textrangle A: 29.2          & O \textrangle A \textrangle C: \textbf{29.2}\\
\bottomrule
\end{tabular}%
}

\caption{The enumerated single prompt CoT against the multi-hop CoT. The table shows the F1 score in percentages across the four datasets and the 3B Qwen and Llama models. The O, C, and A stand for Opinion, Category and Aspect terms accordingly. They show the order of the elements from left to right. Each cell has the top performant element order measured for that dataset and LLM. On the right of the element order is depicted the F1 score, bold shows the higher percentage between Enumerated CoT and Multi-hop CoT.}
\label{tab:cot_vs_multi_hop}
\end{table*}

\end{document}